\documentclass[conference]{IEEEtran}

\makeatletter

\def\ps@IEEEtitlepagestyle{%
  \def\@oddfoot{\mycopyrightnotice}%
  \def\@evenfoot{}%
}
\def\mycopyrightnotice{%
  {\footnotesize 979-8-3503-2781-6      /23/\$31.00 \copyright 2023 European Union\hfill}
  \gdef\mycopyrightnotice{}
}

\usepackage{blindtext}
\usepackage{eso-pic}
\IEEEoverridecommandlockouts
\usepackage{cite}
\usepackage{amsmath,amssymb,amsfonts}
\usepackage{algorithmic}
\usepackage{graphicx}
\usepackage{textcomp}
\usepackage{xcolor}
\def\BibTeX{{\rm B\kern-.05em{\sc i\kern-.025em b}\kern-.08em
    T\kern-.1667em\lower.7ex\hbox{E}\kern-.125emX}}
    
\usepackage{eso-pic}
\newcommand\AtPageUpperMyright[1]{\AtPageUpperLeft{%
 \put(\LenToUnit{0.17\paperwidth},\LenToUnit{-2cm}){%
     \parbox{0.9\textwidth}{\raggedleft\fontsize{8}{11}\selectfont #1}}%
 }}%
\newcommand{\conf}[1]{%
\AddToShipoutPictureBG*{%
\AtPageUpperMyright{#1}
}
}    

\usepackage{tikz}
\usepackage{pgf}
\usepackage{pgfplots}
\usepackage{multirow}
\usepackage{booktabs}

\begin{document}
\title{\vspace*{1cm} Open-Source Tool Based Framework for Automated Performance Evaluation of an AD Function \\
\thanks{*These authors contributed equally to this work.}
}

\author{\IEEEauthorblockN{1\textsuperscript{st} Sanath Konthala*}
\IEEEauthorblockA{\textit{Institute Thermodynamic/Combustion Technology} \\
\textit{FH Aachen - University of Applied Sciences }\\
Aachen, Germany \\
sanath.konthala@alumni.fh-aachen.de}
\and
\IEEEauthorblockN{2\textsuperscript{nd} Daniel Becker*}
\IEEEauthorblockA{\textit{Institute for Automotive Engineering} \\
\textit{RWTH Aachen University}\\
Aachen, Germany \\
daniel.becker@ika.rwth-aachen.de}
\and 
\IEEEauthorblockN{3\textsuperscript{rd} Lutz Eckstein}
\IEEEauthorblockA{\textit{Institute for Automotive Engineering} \\
\textit{RWTH Aachen University}\\
Aachen, Germany \\
lutz.eckstein@ika.rwth-aachen.de}
}

\maketitle
\conf{\textit{  III. International Conference on Electrical, Computer and Energy Technologies (ICECET 2023) \\ 
16-17 November 2023, Cape Town-South Africa}}
\begin{abstract}
As automation in the field of automated driving (AD) progresses, ensuring the safety and functionality of AD functions (ADFs) becomes crucial. Virtual scenario-based testing has emerged as a prevalent method for evaluating these systems, allowing for a wider range of testing environments and reproducibility of results. This approach involves AD-equipped test vehicles operating within predefined scenarios to achieve specific driving objectives. To comprehensively assess the impact of road network properties on the performance of an ADF, varying parameters such as intersection angle, curvature and lane width is essential. However, covering all potential scenarios is impractical, necessitating the identification of feasible parameter ranges and automated generation of corresponding road networks for simulation. Automating the workflow of road network generation, parameter variation, simulation, and evaluation leads to a comprehensive understanding of an ADF's behavior in diverse road network conditions. This paper aims to investigate the influence of road network parameters on the performance of a prototypical ADF through virtual scenario-based testing, ultimately advocating the importance of road topology in assuring safety and reliability of ADFs.
\end{abstract}



\section{Introduction}
The rise of automated driving (AD) has marked a significant milestone in the evolution of the automotive industry, revolutionizing the concept of automobiles. The advent of driverless cars has ushered in a transformative advancement, empowering vehicles to operate autonomously and reducing reliance on human drivers. However, amidst the promises of AD, its impact on traffic safety remains the most crucial aspect. The assertion of accident reduction and improved safety presents a profound opportunity and a daunting challenge for ADFs. Ensuring that these systems surpass human drivers is a complex task. Experts agree that the conventional approach of conducting statistical tests, as done in the automotive industry for previous generations, is not feasible~\cite{LEV11}. Some sources advocate the need for testing over billions of kilometers~\cite{WIN13}, as per International Organization for Standardization (ISO) norm 26262~\cite{ISO2022}, throughout the development lifecycle. 

\begin{figure}[t] 		
	\centering
	\includegraphics[width=\columnwidth]{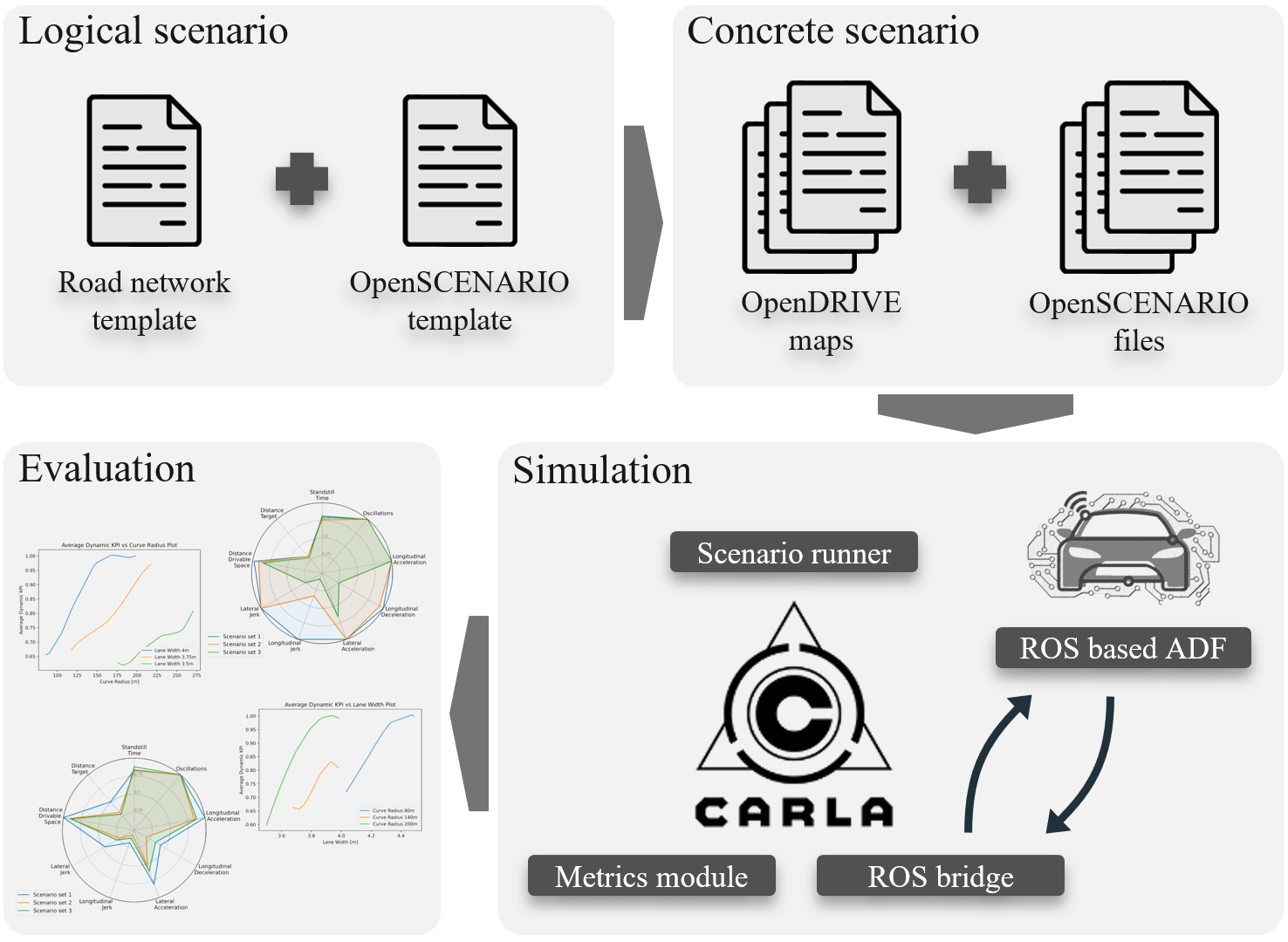}
	\caption{Overview of the implemented automated scenario-based testing methodology. All components and formats are open-source.}
	\label{fig_motivation}
\end{figure}

To address these challenges for assessment of ADFs, virtual scenario-based testing is a promising solution~\cite{ZLO15}. In this approach, an ADF is challenged by creating predefined scenarios that simulate real-world driving conditions. To avoid arbitrary specifications of scenarios, the New Assessment/Test Method (NATM) guidelines propose a structured approach for classifying and describing scenarios based on different levels of abstraction~\cite{VMAD2022}. At the highest level, \textit{functional scenarios} provide a fundamental description of the scenario, including the ego vehicle's actions, interactions with other road users and objects, road geometry and environmental conditions. \textit{Logical scenarios} define value ranges or probability distributions for each element, allowing for a range of possible scenarios. At the most explicit level, \textit{concrete scenarios} are developed by selecting specific values for each element, ensuring a diverse set of test situations and reproducibility. In this work, we perform scenario-based testing in an attempt to understand the influence of geometric parameters alone on an ADF's performance. The overall framework is sketched in Fig.~\ref{fig_motivation}.

The remainder of this paper is structured as follows. First, an overview on related work is given in Sec.~\ref{relatedwork}, followed by an introduction of the open-source tools we utilized in our implementation (cf. Sec.~\ref{researchapproach}). In Sec.~\ref{methodology}, the overall methodology is described in which we first define logical scenarios by specifying value ranges for the varied parameters. We then generate sets of concrete scenarios. Subsequently, the ADF is tested on each scenario, and its performance is evaluated using a set of Key Performance Indicators (KPIs). After that, some results are discussed in Sec.~\ref{sec_results} and finally, we draw a conclusion and give an outlook (cf. Sec.~\ref{sec_conc}).
 
\section{Related Work} \label{relatedwork}
This chapter provides an overview of the existing research related to the assessment of ADFs and the underlying motivation behind this study.

\subsection{Scenario-based Testing}\label{SBT}
Virtual scenario-based testing offers a promising alternative to real-world testing for ADFs, enabling the evaluation of driving functions under diverse situations by varying scenario parameters. However, a significant challenge, known as “parameter space explosion” arises due to the large number of parameters defining a scenario, leading to an exponentially increasing number of possible parameter combinations~\cite{AME19}. As a result, it becomes impractical to include all theoretically possible scenarios in the safety validation process. To address this, \cite{AME19} proposes a method to define the required test coverage for scenario-based validation. In~\cite{RIE20}, a comprehensive literature review on scenario-based testing for ADF is conducted. They explored various approaches for defining and selecting scenarios to achieve efficient testing and conducted a comparative analysis of different methods. According to the taxonomy of the scenario-based testing introduced by them, the source of the scenarios is from abstract knowledge of experts, like German guidelines for the construction of highways~\cite{RAA08} or from existing driving and accident data. 

\subsection{Scenario Contents}
\begin{figure}[htb] 		
	\centering
	\includegraphics[width=0.65\columnwidth]{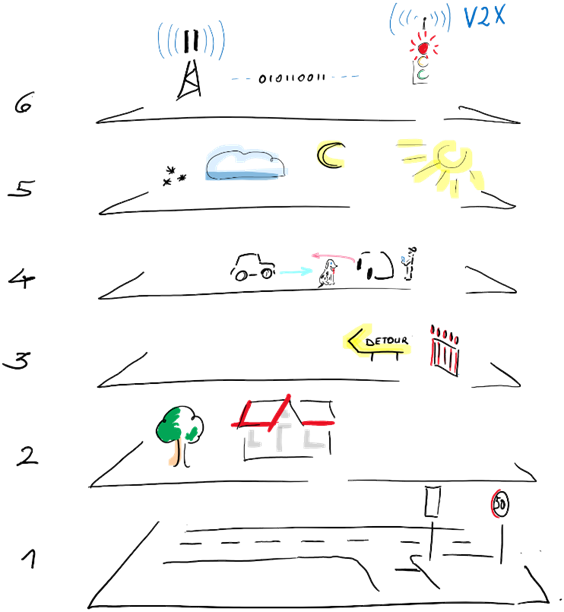}
	\caption{The 6 layer model to structure a scenario according to~\cite{SCH21}.}
	\label{fig_6LM}
\end{figure}

The scenario description in scenario-based testing is critical to ensure comprehensive testing and proper execution. The 6 layer model (6LM) shown in Fig.~\ref{fig_6LM} offers a structured approach with six layers for organizing essential elements of a scenario~\cite{SCH21}. Layer 1 focuses on permanent objects required for traffic guidance, such as road network related elements. Layer~2 includes non-road objects like buildings and vegetation. Moving to Layer~3, temporary objects from layers~1 and 2 are described, such as road work signs or fallen trees. Layer~4 introduces time-dependent description, covering dynamic objects that move or can potentially move. This layer includes traffic participants, parked cars, and stationary pedestrians. In Layer~5, environmental conditions like weather and lighting are accounted for, with effects that may change over time. Layer~6 incorporates digital data, including information exchange, communication, and traffic light status. Intelligent traffic management systems using V2X and V2V communication or cellular networks fall into this layer.

 Existing research on safety assurance for ADFs has primarily focused on layers 3, 4, and 5 of the 6LM~\cite{RIE20, WAG21}. However, safety assurance concerning layer 1, i.e., road network geometry, has been lacking, leaving gap in understanding the impact of road geometry design on driving function performance and safety. In \cite{TAK22} an analysis of the relationship between road geometry and automated driving safety for vehicle-based mobility services is conducted. Their study was limited to simulations and evaluation based on a real-time dataset, lacking coverage of a wide range of scenarios. However, the study highlights the importance of considering road network variations to simulate ADFs and evaluate their performance, ultimately ensuring safety assurance of the driving function. Building upon the work in \cite{BEC23} and \cite{beckerRoadVar}, this paper extends the approach for performance analysis of ADFs with the use of open-source tools. The study aims to test the ADF in a more complex and diverse environment, further exploring its capabilities and behavior under varied road network conditions. 
 
\section{Research Approach}\label{researchapproach}
The core idea of this work is to use open-source tools and formats for evaluating the ADF's performance. This section introduces the existing standards and open-source tools that have been employed to illustrate the research approach of the study. 
 
\subsection{Road and Scenario Description Standards}\label{RSDS}
In this section, a brief overview of existing road and scenario description formats is given.
 
\subsubsection{OpenDRIVE}
ASAM OpenDRIVE is an open standard utilized for detailing road networks and topologies\footnote{https://www.asam.net/standards/detail/opendrive/}.
It employs XML syntax with the file extension ".xodr", providing a standardization foundation for describing road networks. The OpenDRIVE file stores various data elements to outline road and lane geometries, road marks, and other objects like signals. These road networks can be artificially generated or based on actual data. While the OpenDRIVE format primarily serves simulations within simulators and allows for rapid map switching, it poses challenges in interacting with ADFs. 

\subsubsection{Lanelet2}
The Lanelet2 framework contains a C++ library and a map format written in XML~\cite{POG18}. It has been gaining popularity for research on ADFs, as it provides comprehensive functionality and meets the requirements of both driving simulators and AD tasks. A Lanelet2 map is an extension of OpenStreetMap, featuring three layers: the physical layer with observable elements, the relational layer connecting them to lanes and traffic rules, and the topological layer forming a network of passable regions. The Lanelet2 library allows seamless interaction with Lanelet2 files, enabling various operations like processing and manipulation of map data for AD applications.

\subsubsection{OpenSCENARIO}
ASAM OpenSCENARIO is a standardized format and methodology for describing scenarios in driving and traffic simulators to test ADFs\footnote{https://www.asam.net/standards/detail/openscenario/}.
It enables comprehensive descriptions of complex maneuvers involving multiple elements like vehicles and pedestrians. Scenarios can be derived from driver actions or recorded driving maneuvers, precisely defining the dynamic content of the simulated world, including traffic participants' behavior. The format introduces a hierarchical structure for scenario descriptions, enabling the creation of storyboards, stories, acts, and maneuvers. While focusing on defining the dynamic content, it can reference and integrate road network descriptions from ASAM OpenDRIVE, providing a comprehensive framework for describing both static and dynamic aspects of simulation applications.

\subsection{Open-Source Tools and Software}
This section provides a brief introduction to the essential tools and software utilized in this study.

\subsubsection{Road Generation Tool}
The Road Generation Tool is an open-source tool that simplifies the creation of OpenDRIVE road networks using XML-based more logical and less redundant descriptions than OpenDRIVE~\cite{becker2020roadgen}. Users can specify parameters and values in an XML template to generate standardized simulation maps in the OpenDRIVE format. The tool provides command-line functionality for creating various road geometries, such as junctions and intersections. It also includes a Python package that allows the incorporation of stochastic variables for generating multiple variations of the road network. 
 
\subsubsection{CommonRoad Scenario Designer}
The CommonRoad Scenario Designer~\cite{MAI21} is a versatile toolbox for creating, manipulating, and converting road maps and scenarios. It supports multiple map formats, including Lanelet2, OpenDRIVE, OpenStreetMap (OSM), and SUMO. The toolbox offers a graphical user interface (GUI), command line interface, and Python APIs to facilitate the creation, editing, and visualization of CommonRoad maps and scenarios. One of its key features is the ability to convert maps between different formats, such as OpenDRIVE to Lanelet2 format.

\subsubsection{Robot Operating System (ROS)}
ROS is an open-source meta-operating system\footnote{https://www.ros.org/}
used for developing and testing robotic systems, including ADF validation. Its benefits include modularity, seamless communication, visualization and debugging tools, a vibrant community, and integration with simulation environments. ROS provides a flexible and collaborative framework for developing, integrating, and testing components in ADF.

\subsubsection{CARLA Simulator}
CARLA is an open-source simulation platform~\cite{DOS17} designed for developing, training, and validating AD systems. It offers open digital assets like urban layouts, buildings, and vehicles, allowing flexible specification of sensor suites, environmental conditions, and full control over static and dynamic actors. The platform generates realistic and high-fidelity virtual environments, providing users with a space to test and evaluate their algorithms and systems. CARLA’s extensive API enables users to control the simulation, access sensor data, and interact with the environment. 
The different CARLA modules used in this study are the ROS bridge, ScenarioRunner and Metrics Module. The former establishes seamless communication between ROS topics and CARLA, enabling two-way data exchange. This way, an ADF can interact effectively with the simulation environment. The ScenarioRunner acts as the scenario engine within the simulation and can process OpenSCENARIO files and orchestrates the various entities of a given scenario. Finally, the metrics module is used for calculations and monitoring of parameters after a scenario is finished in the CARLA simulator. The recorded simulation information is used as input, and customized metrics can be defined.

\section{Methodology}\label{methodology}
To comprehend the influence of the road network geometry on the performance of the ADF, it is crucial to conduct tests across various scenarios. For this purpose, an automated framework has been created that facilitates efficient scenario generation and simulation. The workflow is visualized in Fig.~\ref{fig_motivation} and consists of three main components: the \textit{automated scenario generation} of logical and concrete instances, an \textit{automated simulation framework}, and the \textit{simulation evaluation}. This streamlined process allows to systematically explore and assess the impact of road network geometry on the performance of the ADF and will be detailed in the following sections. 

\subsection{Automated Scenario Generation}\label{sec_scenario_gen}
To assess the ADF's performance across diverse road geometry parameters, two basic road topology templates and a complex road network are utilized. The variation tool within the road generation tool is employed to facilitate the creation of various iterations. Different scenarios are generated using a linear distribution approach for the parameter values, allowing the efficient creation of a range of varied scenarios and evaluation of the ADF's behavior under different road geometries.

For each template, logical scenarios are initially defined by incorporating various variables that defined the road geometry. These variables represent different aspects of the road design. Each logical scenario consists of a road network template and an OpenSCENARIO file containing initial configuration settings for co-simulation, which encompass the ego vehicle's start and destination locations. The OpenSCENARIO file also contains essential simulation details, such as vehicle information, scenario specifics, and environment data. As most settings in the OpenSCENARIO file are default and remain unchanged, this logical OpenSCENARIO file is used in the simulation framework to generate concrete OpenSCENARIO files. In addition, The variation tool is used to generate concrete OpenDRIVE maps. All templates consist of roads with two lanes — one for each direction. It is worth noting that elevation has not been taken into account while creating scenarios for this particular study.

The simulation platform requires an OpenDRIVE map, while the AD platform relies on a lanelet2 map. Consequently, each OpenDRIVE map is converted to a lanelet2 map with help of the CommonRoad scenario designer. Then, both maps and the OpenSCENARIO file are stored to form a scenario database. To efficiently handle the large number of generated scenarios, an automated approach has been implemented for scenario generation and lanelet2 conversion using Python scripts. This automated workflow streamlines the process of scenario creation and ensures the availability of the required files for simulation execution. The different parameters used to define the logical scenarios are summarized in Tab.~\ref{tab_param}. 

\begin{figure*}[t]
\centering
	\includegraphics[width=\textwidth]{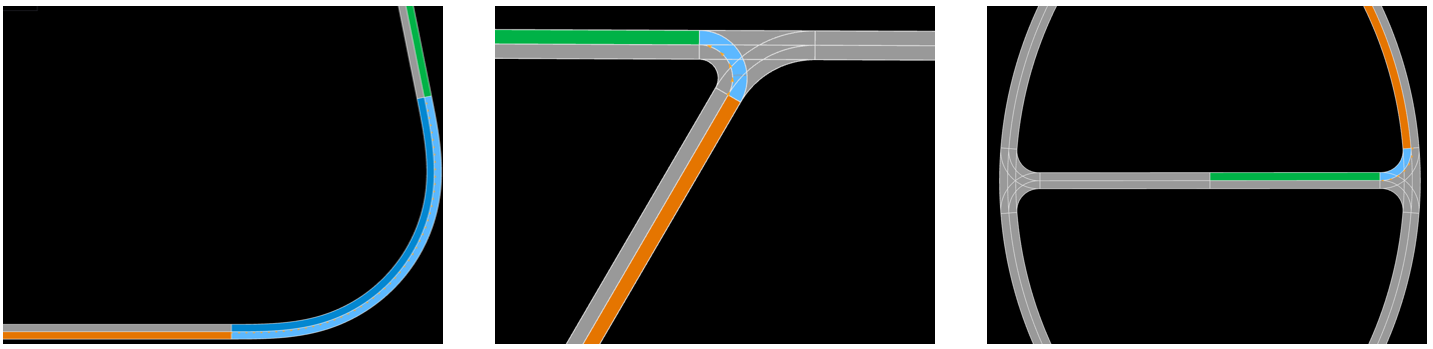}
	\caption{Concrete examples of the three utilized road network templates. Exemplary routes are drawn in each scenario.}
	\label{fig_templates}
\end{figure*}

The description of the individual templates used in this study are given as follows and examples for each template are visualized in Fig.~\ref{fig_templates}.

\begin{table}[htb]
	\caption{Parameter spaces and combinations for the road network generation resulting in a total of 8 logical scenarios.}
	\label{tab_param}
	\begin{center}
		\begin{tabular}{lll}
			\textbf{Scenario} &\textbf{Varied parameter} & \textbf{Constant param.} \\
			\toprule
			Curved road (left) & Lane width: $3.4$ - $4.5\,\text{m}$ & Curve radius\\
            \midrule
			Curved road (right) & Lane width: $3.0$ - $4.0\,\text{m}$ & Curve radius\\
            \midrule
			Curved road (l+r) & Lane width: $50$ - $500\,\text{m}$ & Lane width\\
            \midrule
			\multirow{3}{*}{T-junction} & Junction gap: $5$ - $30\,\text{m}$ & Intersection angle\\
            & Intersec. angle: $35$ - $135^{\circ}$ & Junction gap\\
            & Lane width: $3.5$ - $4.2\,\text{m}$ & Intersection angle\\
            \midrule
			Complex road & Lane width: $3.5$ - $4.0\,\text{m}$ & curve radius\\
		\end{tabular}
	\end{center}
\end{table}

\subsubsection{Curved Road}
The first template employed in this research is a curved road, which serves as a fundamental exploration of road network parameters like lane width and curve radius. To examine the impact of the curve's direction, both left-turning and right-turning curve geometries were utilized. 

\subsubsection{T-junction}
The second template used in this study represents a T-junction, which serves as a simulation of a road junction area. The parameters varied under this template include the intersection angle, lane width and junction gap. The latter describes the size of the actual junction measured from the virtual crossing point of all junction arms. The primary objective is to gain deeper insights into low speed turning maneuvers at the intersection. To achieve this, simulations are conducted for both left-turning and right-turning scenarios to comprehensively assess the ADF's behavior in diverse situations.


\subsubsection{Complex Road}
A combination of a T-junctions and a curved road is used to form a complex road network. The main goal of this is to test the ADF under challenging road conditions. 

\subsection{Automated Simulation Framework}
Once the scenario database is established, the subsequent stage entails selecting individual scenarios from the database to create a concrete test case for the automation framework. For each logical scenario, to create corresponding concrete simulation scenarios, the template OpenSCENARIO file with default settings is imported, and then scenario-specific information (e.g., scenario name) is added to the template to generate the desired concrete OpenSCENARIO file. The road name in the OpenSCENARIO file indicates the OpenDRIVE map required for simulation. The start road, lane ID, and position define the initial location where the ego vehicle is spawned, while the target road, lane ID, and position determine the destination on the OpenDRIVE map. To streamline this process automatically, a Python script is used to initialize simulation settings. This script is seamlessly integrated into the simulation pipeline, ensuring efficient and consistent configuration of test cases for evaluation.

The utilized system-under-test for conducting simulations is a prototypical ADF based on \cite{Kueppers2022}. As the "Simulation" block of Fig.~\ref{fig_motivation} shows it is connected via ROS and integrated in a closed-loop manner with the previously described CARLA ROS bridge. An automated simulation framework has been designed to efficiently run scenarios from the scenario database. The process involves selecting a scenario, generating its corresponding OpenSCENARIO file, and executing the simulation facilitated by CARLA ScenarioRunner. A successful simulation occurs when the ego vehicle reaches its destination and comes to a stop. To avoid endless attempts for failed scenarios, a predetermined limit is set on the number of simulation attempts. Simulation data is recorded and stored in the results database. After completing all simulations, the CARLA metrics module is employed to extract essential data for KPI calculation from the recorded files to an independent format, such as CSV. The automation framework is developed in a highly generic manner to enable easy utilization for users with their ADF. Minor adjustments to system-specific parameters allow seamless integration, making the framework accessible to a wide range of users.

\subsection{Simulation Evaluation}\label{sim_eval}
After completing all simulations, the next step involves analyzing the data to evaluate the ADF's performance. Various KPIs have been computed from the extracted vehicle data, including Longitudinal Acceleration, Longitudinal Deceleration, Lateral Acceleration, Longitudinal Jerk, Lateral Jerk, and Distance Target. These metrics are normalized using a reference value within the range of [$0$,~$1$]. A KPI value of $0$ signifies very poor performance, while a value of $1$ indicates good performance. This normalization process ensures a standardized evaluation across all criteria, providing a clear and consistent understanding of the ADF's performance level, ranging from poor to good.

The reference values used for evaluating the vehicle's longitudinal motion are based on the performance requirements outlined in the ACC system of ISO15622~\cite{ISO09}. According to these requirements, the reference value for longitudinal acceleration is established at $2\,\text{m}/\text{s}^2$, and the reference value for longitudinal deceleration is set at $3.5\,\text{m}/\text{s}^2$. Additionally, the reference value for vehicle lateral acceleration is defined as $3\,\text{m}/\text{s}^2$, while for longitudinal and lateral jerk, it is set at $5\,\text{m}/\text{s}^3$. These reference values serve as benchmarks to assess the ADF's performance in longitudinal and lateral motions, ensuring adherence to standardized criteria. These KPIs are also referred to as dynamic KPIs in further sections. 

Additionally, comfort-based analysis was conducted on the ADF to address surge and oscillations in motion planning and control systems. Surge refers to engine power variations under steady throttle, causing speed fluctuations during straight-line driving, while oscillations in specific frequency ranges (e.g., $1$ to $2\,\text{Hz}$) can affect passenger comfort. Vroom's comfort analysis approach~\cite{VRO21} uses the Root Mean Square (RMS) value of acceleration based on ISO2631-1:1997~\cite{ISO97} guidelines to evaluate comfort. It presents acceleration RMS values based on an 8-hour exposure as a comfort reference as shown in Tab.~\ref{tab_comfort}. Vertical vibrations were not considered in this study due to the absence of road height profiles.

\begin{table}[ht]
	\caption{Comfort perception to vibrating environments taken from~\cite{ISO97} used to assess the comfort level.}
	\label{tab_comfort}
	\begin{center}
		\begin{tabular}{cl}
			\textbf{Acceleration (x-direction) [$m/s^2$]} & \textbf{Perceived comfort} \\
			\toprule
			$\le 0.314$ & not uncomfortable \\
            \midrule
            $0.315 \text{ to } 0.63$ & a little uncomfortable \\
            \midrule
            $0.5 \text{ to } 1.0$ & fairly uncomfortable \\
            \midrule
            $0.8 \text{ to } 1.6$ & uncomfortable \\
            \midrule
            $1.25 \text{ to } 2.5$ & very uncomfortable \\
            \midrule
            $\ge 2.0$ & extremely uncomfortable			
		\end{tabular}
	\end{center}
\end{table}

The acceleration RMS value represents the average longitudinal acceleration over a specific time period. It is calculated using the variable $\Tilde{a}$, which is the weighted vehicle acceleration in $\text{m}/\text{s}^2$ as shown in (\ref{eq_rmsComfort}). The acceleration is filtered through a band-pass filter with a bandwidth of $1$ to $32$\,Hz. The starting time is denoted as ${t_0}$ and ${t_f}$ represents the final time.
\begin{equation}
	RMS = \sqrt{\frac{\int_{t_f}^{t_0} \Tilde{a}^2\,dt}{t_f - t_0} }
	\label{eq_rmsComfort}
\end{equation}

\section{Results}\label{sec_results}
\begin{figure}[htb] 		
	\centering
    \input{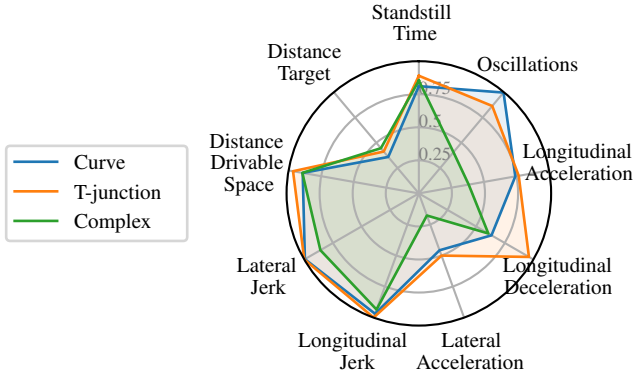}
	\caption{Illustration of the overall performance evaluation for all executed simulations clustered by the three main templates. The spider plot indicates that the ADF generally have good jerk values and almost always stays within the lane boundaries. Reaching the target seems to be an issue which might be controller related since the position is overshot. In the complex road network, acceleration and steady lane keeping values tend to be bad which may be due to the transitions between different road elements.}
	\label{fig_spider}
\end{figure}

Fig. \ref{fig_spider} shows the overall performance of the ADF on $1460$ variations for three templates used in this study. The average KPIs of curved road, T-junction, and complex road are $0.737$, $0.816$, and $0.605$ respectively. In the case of the curved road scenario, the vehicle encountered difficulties at certain curve radii for various lane widths present in the database. However, the ADF demonstrated the ability to consistently maintain the lane center throughout the journey. In the T-junction scenario, the vehicle efficiently reached a steady state and decelerated in time before approaching the turn at the intersection. This resulted in stable performance, with the vehicle maintaining the lane center and demonstrating good overall dynamic performance. The incorporation of curved road and T-junction in the complex road template led to numerous crashes, especially in sharp curves and narrow lane widths. These crashes directly contributed to the poor overall performance observed in the KPI plot. In this section, the  analysis of the three templates used in this study for the evaluation of the ADF will be discussed in more detail.

\subsection{Curved Road}
\begin{figure}[htb] 		
	\centering
    \input{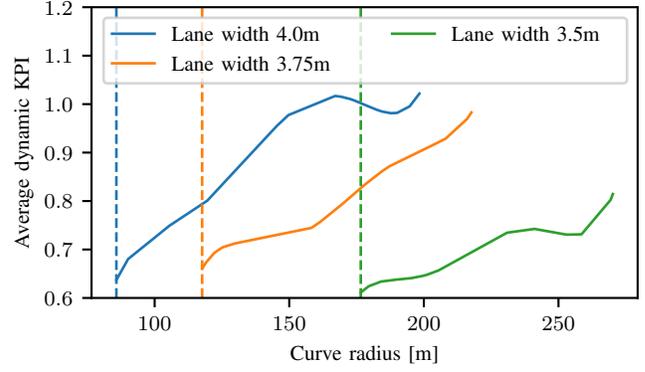}
	\caption{Filtered average of the dynamic KPIs in left turns with three fixed lane width over different curve radii. For each lane width, a critical radius was identified below which the ADF failed to succeed (dashed vertical lines). Expectable, the performance in wider lanes seems to  be better.}
	\label{fig_curveDyn}
\end{figure}

The curved road template includes both left and right turning scenarios. For each set of variation in both scenarios, the defined KPIs are calculated.The average KPI value for left-turning scenarios was found to be $0.665$, while for right-turning scenarios, it was $0.832$. Remarkably, the performance witnessed a substantial $20\%$ decline during the transition from right-turning to left-turning simulations. Furthermore, through failed simulations, a critical radius has been identified for left turning scenarios. This critical radius represents the curve's radius below which the ADF failed to accomplish the specified objective. Overall, $6$ simulations failed in left turning scenario and critical radii of approximately $86\,\text{m}$, $118\,\text{m}$, and $177\,\text{m}$ for lane width of $4.0\,\text{m}$, $3.75\,\text{m}$, and $3.5\,\text{m}$ respectively. On the other hand, all simulations were successful in right turning scenario. The dynamic KPI (cf. Sec.~\ref{sim_eval}) values for three fixed lane widths and varied curve radii in left curves is shown in Fig.~\ref{fig_curveDyn}. The improved performance with respect to the curve radius can be clearly seen. The data indicate a positive trend in dynamic KPI as the curve radius increases in each lane width scenario. In addition, the performance tends to be better with wider lane width. 

The oscillation KPI shown in Fig.~\ref{fig_spider} can be interpreted such that in a continuous drive through curves, the ADF maintains a stable and steady driving trajectory.

\subsection{T-junction}
\begin{figure}[htb] 		
	\centering
    \input{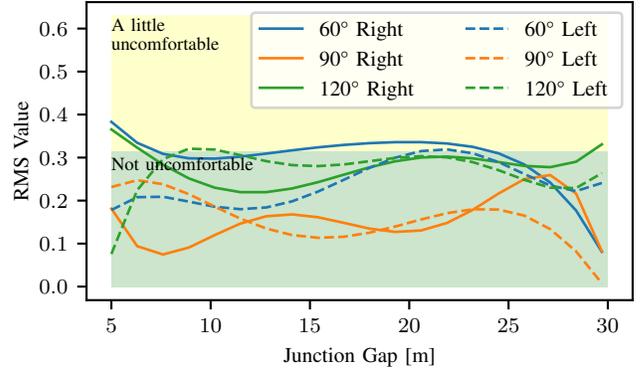}
	\caption{Average comfort levels with respect to longitudinal acceleration of three fixed intersection angles. The boundaries are based on Tab.~\ref{tab_comfort}. A minor trend to more comfort with increasing junction gap can be noted. Also, left turns seem to result in more comfort which might be due to the larger radius on the junction.}
	\label{fig_curveComfort}
\end{figure}

Overall, the ADF performed well on variations of T-junction. Due to the presence of low speed turning maneuver, the KPI values are observed in good region. To understand the comfort throughout the journey, mean RMS value of longitudinal acceleration has been calculated and is shown in Fig.~\ref{fig_curveComfort} which are based on the study results from Tab.~\ref{tab_comfort}. The overlaps are estimated towards the worse value, respectively. The plot indicates that with increase in junction gap, the comfort perception also increases. Due to the long turning section based on the direction of the turn, good performance was observed at 60° in left turning scenario and 120° in right turning scenario. 

When looking at the longitudinal deceleration KPI shown in Fig.~\ref{fig_spider} it can be seen that the ADF tends to smoothly decelerate when approaching and passing an intersection. However, the lateral acceleration has room for improvement.

\addtolength{\textheight}{-2.20cm}   

 
\subsection{Complex Road}
The complex road, being a combination of curved road and T-junction, significantly affects the performance of the ADF. The presence of sharp curves at the T-junction and curve section leads to pronounced lateral and longitudinal jerk, impacted the overall performance of the ADF. Additionally, the vehicle's behavior of slowing down at curves and rapidly accelerating contributed to lower overall performance in longitudinal and lateral acceleration KPIs. The overall KPI plot of all three templates used in this study is shown in Fig.~\ref{fig_spider}, indicating that the combination of multiple concatenated road geometry affects the ADF's performance. Also, these transitions between junctions and connection roads might be the reason for the multiple occurring oscillations throughout the scenarior runs.
 
\section{Conclusion and Outlook}\label{sec_conc}
In this research, scenario-based testing was implemented to study the influence of road geometry parameters on the performance of an automated driving function. The evaluation revealed that the ADF performed better in right-turning scenarios compared to left-turning scenarios in the curved road template. The T-junction template exhibited excellent performance in both left and right-turning scenarios. However, challenges arose in the complex road template, resulting in crashes, especially in sharp curves and narrow lanes. These findings confirm that road geometry significantly influences the ADF's performance, aligning with the primary objective of this study.

Looking ahead, further research can focus on enhancing the road network generation tools to better handle complex road scenarios with lane widening and additional lanes. Incorporating road height parameters will also enhance the realism of simulations for urban traffic scenarios with ramps and varying elevations. Increasing the number of generated road networks and exploring multiple variables through machine learning techniques can lead to more diverse and comprehensive scenario variations. 

\section*{Acknowledgment}
The work of this paper has been done in the context of the SUNRISE project which is co-funded by the European Commission’s Horizon Europe Research and Innovation Programme under grant agreement number 101069573. We would like to express our sincere gratitude to Professor Dr.-Ing. Thomas Esch for his invaluable support and guidance in publishing this paper.

Views and opinions expressed, are those of the author(s) only and do not necessarily reflect those of the European Union or the European Climate, Infrastructure and Environment Executive Agency (CINEA). Neither the European Union nor the granting authority can be held responsible for them.






\bibliographystyle{IEEEtran}
\bibliography{ma_sanath}


\end{document}